\documentclass{article} % For LaTeX2e
\usepackage{iclr2026_conference,times}

\usepackage{amsmath,amsfonts,bm}

\def\eqref#1{equation~\ref{#1}}
\def\1{\bm{1}}

\DeclareMathAlphabet{\mathsfit}{\encodingdefault}{\sfdefault}{m}{sl}
\SetMathAlphabet{\mathsfit}{bold}{\encodingdefault}{\sfdefault}{bx}{n}

\usepackage{hyperref}
\usepackage{url}

\usepackage{hyperref}
\usepackage{url}
\usepackage{graphicx}
\usepackage{amsmath}
\usepackage{amssymb}
\usepackage{booktabs}
\usepackage{enumitem}
\usepackage{indentfirst}
\usepackage{multirow}
\usepackage{pifont}
\usepackage{selectp}
\usepackage{makecell}
\usepackage{diagbox}
\usepackage{rotating}
\usepackage{float}
\usepackage{lipsum}
\usepackage{tabularray}
\usepackage{subcaption}
\usepackage{hyperref}  % It's a good practice to load hyperref before cleveref
\usepackage{cleveref}
\usepackage{booktabs} % For prettier table lines
\usepackage{adjustbox} % For table adjustments
\usepackage{diagbox} 
\usepackage{algorithm}
\usepackage{algpseudocode}
\usepackage{amsmath}
\usepackage{tcolorbox}
\tcbuselibrary{skins} 

\newtcolorbox[auto counter, number within=section]{Genprompts}[1][]{
  colback=green!5!white,    % 更淡的背景色
  colframe=green!70!black,  % 更深的边框颜色
  coltext=black,           % 文本颜色保持黑色
  boxrule=0.75pt,          % 边框厚度
  arc=4pt,                 % 更柔和的弧度
  boxsep=4pt,              % 增加边距以更好的排版
  fonttitle=\bfseries,     % 标题加粗
  title=Generation Prompts,
  coltitle=black,          % 标题文本颜色
  drop shadow=black!30,    % 添加阴影效果
  enhanced,                % 增强样式
  #1                       % 允许用户传递额外的参数
}

\usepackage{booktabs}
\usepackage[table]{xcolor}
\usepackage{multirow}
\usepackage{amsmath}
\usepackage{graphicx}

\newcommand{\sota}[1]{\textbf{#1}}

\newcommand{\annotate}[3]{%
    #1\raisebox{-0.5ex}{\scriptsize\textcolor{#2}{#3}}%
}

\usepackage{wrapfig}
\usepackage{booktabs} 
\usepackage{hyperref}
\usepackage{url}

\title{DRQA: Dynamic Reasoning Quota Allocation for Controlling Overthinking in Reasoning Large Language Models}
\author{
 \textbf{Kaiwen Yan\textsuperscript{1}},
 \textbf{Xuanqing Shi\textsuperscript{3}},
 \textbf{Hongcheng Guo\textsuperscript{4}}, \\
 \textbf{Wenxuan Wang\textsuperscript{5}}, 
 \textbf{Zhuosheng Zhang\textsuperscript{6}},
 \textbf{Chengwei Qin\textsuperscript{1,2}}
\\
\\
 \textsuperscript{1}The Hong Kong University of Science and Technology (Guangzhou),\\
 \textsuperscript{2}The Hong Kong University of Science and Technology,
 \textsuperscript{3}Tsinghua University,\\
 \textsuperscript{4}Fudan University,
 \textsuperscript{5}Renmin University of China,
 \textsuperscript{6}Shanghai Jiao Tong University
}

\iclrfinalcopy
\begin{document}

\maketitle
\begin{abstract}
Reasoning large language models (RLLMs), such as OpenAI-O3 and DeepSeek-R1, have recently demonstrated remarkable capabilities by performing structured and multi-step reasoning. However, recent studies reveal that RLLMs often suffer from overthinking, i.e., producing unnecessarily lengthy reasoning chains even for simple questions, leading to excessive token consumption and computational inefficiency. Interestingly, we observe that when processing multiple questions in batch mode, RLLMs exhibit more resource-efficient behavior by dynamically compressing reasoning steps for easier problems, due to implicit resource competition. Inspired by this, we propose \emph{Dynamic Reasoning Quota Allocation (DRQA)}, a novel method that transfers the benefits of resource competition from batch processing to single-question inference. Specifically, DRQA leverages batch-generated preference data and reinforcement learning to train the model to allocate reasoning resources adaptively. By encouraging the model to internalize a preference for responses that are both accurate and concise, DRQA enables it to generate concise answers for simple questions while retaining sufficient reasoning depth for more challenging ones. Extensive experiments on a wide range of mathematical and scientific reasoning benchmarks demonstrate that DRQA significantly reduces token usage while maintaining, and in many cases improving, answer accuracy. By effectively mitigating the overthinking problem, DRQA offers a promising direction for more efficient and scalable deployment of RLLMs, and we hope it inspires further exploration into fine-grained control of reasoning behaviors.
\end{abstract}

\section{Introduction}
Reasoning large language models (RLLMs), such as OpenAI-O3 ~\citep{openaio3} and DeepSeek-R1 ~\citep{deepseekai2025deepseekr1incentivizingreasoningcapability}, have recently showcased remarkable capabilities in complex problem solving and decision-making, achieving state-of-the-art performance across a wide range of tasks. However, recent studies have revealed that LLMs often generate unnecessarily lengthy reasoning chains, even for simple questions like ``2+3=?''~\citep{sui2025stopoverthinkingsurveyefficient, chen2025think23overthinkingo1like}. While extended reasoning can improve accuracy on complex tasks, this tendency to \emph{overthink} leads to excessive token usage and growing computational and economic costs, posing significant challenges for the scalable and practical deployment of RLLMs in real-world scenarios.

\begin{figure*}
    \centering
    \includegraphics[width=0.9\textwidth]{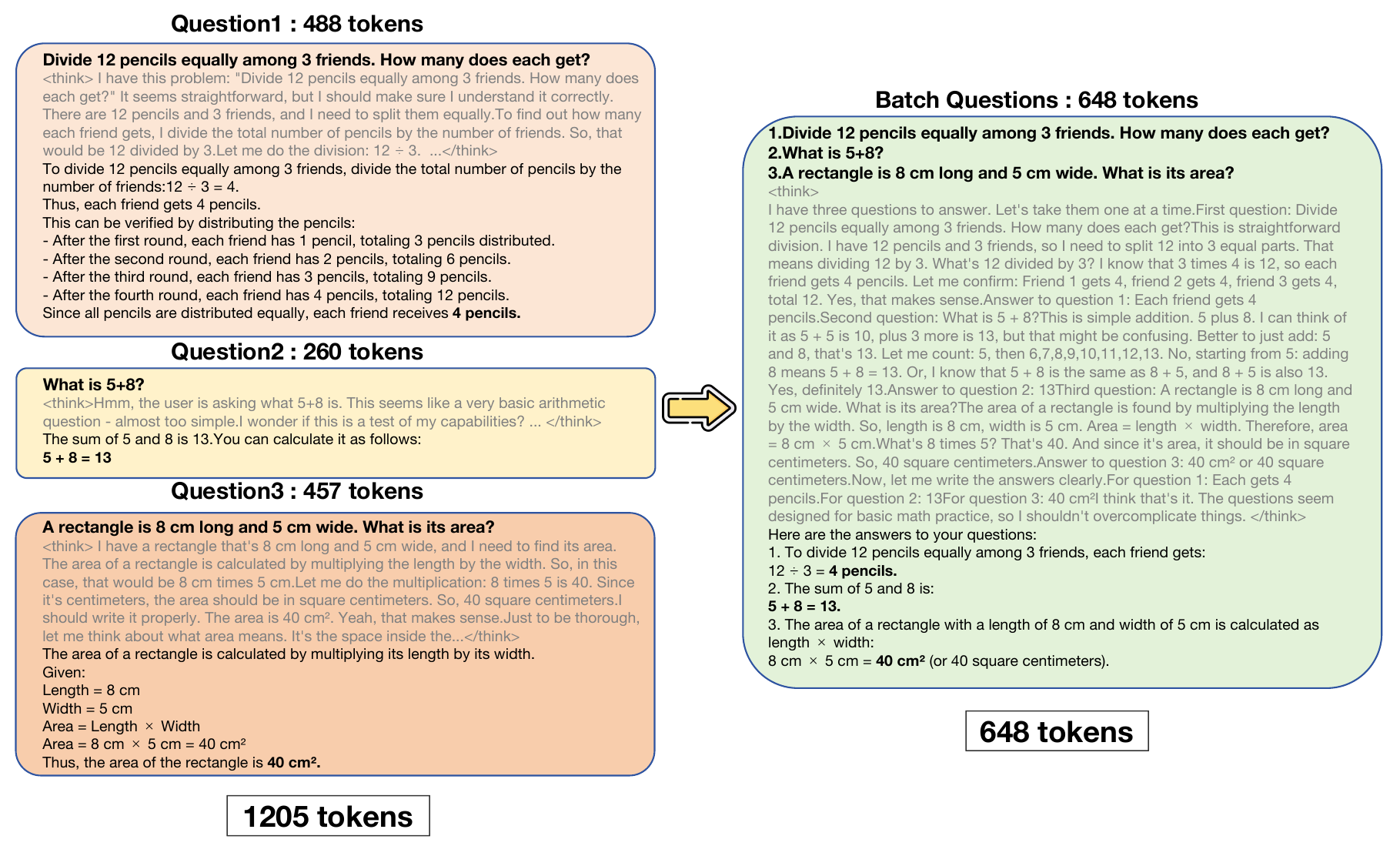}
    \caption{Comparison between batch inference and single-question inference using Deepseek-R1. Answering three questions together results in significantly fewer tokens than answering each question individually.}
    % \vspace{-1.2em}
    \label{fig:batch_example}
\end{figure*}
% Interestingly, we uncover an emergent efficiency behavior during \emph{batch inference}: when multiple questions are processed together, the total output length is significantly shorter than the combined length of the responses generated for each question individually.
Inspired by recent findings on instruction-tuned LLMs~\citep{lin2024batchpromptaccomplish,cheng2023batchpromptingefficientinference}, which show that processing multiple inputs together during \emph{batch inference} can reduce the total generated length compared to answering them individually, we investigate whether a similar phenomenon exists in RLLMs. Our study reveals that this effect in RLLMs goes beyond mere solution shortening: batch inference also \emph{compresses the chain-of-thought reasoning process} itself.
For example, as shown in Figure~\ref{fig:batch_example}, answering three questions together yields only 648 tokens in total, compared to 1205 tokens when answered separately. This suggests that under a shared context window, questions implicitly compete for a global reasoning quota, prompting the model to prioritize essential logic and suppress redundancy, an effect we refer to as ``\emph{resource competition pressure}''.

This observation raises a core research question: can the benefits of resource competition in batch inference be transferred to single-question settings? If so, RLLMs could dynamically adjust their reasoning behaviors, offering concise responses for simple questions while allocating more resources to more complex ones. To this end, we introduce \textbf{D}ynamic \textbf{R}easoning \textbf{Q}uota \textbf{A}llocation (\textbf{DRQA}), a novel approach that brings the advantages of resource competition into single-question inference, enabling more efficient and adaptive reasoning. Specifically, we first collect diverse reasoning chains under batch inference settings and analyze how the model automatically allocates the length of reasoning chains to problems of varying difficulty in the presence of resource competition. We then construct a preference dataset and introduce a reinforcement learning objective that enables the model to distinguish and learn the advantages of ``concise and accurate'' reasoning chains over those that are ``verbose or incorrect''. By indirectly encouraging the model to favor the ``concise and accurate'' patterns that emerge from batch inference, we enhance its overall reasoning capabilities.

We evaluate the effectiveness of DRQA across a diverse set of reasoning benchmarks, including GSM8K~\citep{cobbe2021trainingverifierssolvemathgsm8k}, MATH-500~\citep{math500hendrycks2021measuringmathematicalproblemsolving}, AIME 2024 and 2025~\citep{aime}, AMC~\citep{AMC2023}, GPQA-Diamond~\citep{rein2023gpqagraduatelevelgoogleproofqa} and LiveCodeBench~\citep{jain2024livecodebenchholisticcontaminationfree}. Experimental results show that DRQA reduces token usage by over \textbf{30\%} while consistently maintaining or improving answer accuracy, offering an effective and scalable solution to the overthinking problem. In summary, our main contributions are:
\begin{itemize}[leftmargin=*,topsep=2pt,itemsep=2pt,parsep=0pt]
    \item To the best of our knowledge, we for the first time systematically investigate how \emph{``resource competition pressure''} can enhance the reasoning efficiency of RLLMs during batch inference.
    \item We propose DRQA, a novel method that transfers this efficiency mechanism to single-question inference by leveraging batch-generated preference data and reinforcement learning. This enables the model to generate \emph{concise answers for simple questions} while maintaining \emph{deep reasoning for complex ones}.
    \item With extensive experiments, we demonstrate the effectiveness of DRQA compared to existing ones and analyze the results thoroughly.
\end{itemize}

\section{Resource Competition During Batch Inference}

% \begin{table}[htbp]
% \caption{Comparison of average output token lengths across different models under the `Vanilla' and `Batch-2' settings.}
% \label{tab:batch_compression}
% \centering
% \resizebox{0.4\textwidth}{!}{
% \begin{tabular}{lcc}
%     \toprule
%     \textbf{Model} & \textbf{Vanilla} & \textbf{Batch-2} \\
%     \midrule
%     Deepseek-R1           & 5640.4  & 4035.2 \\
%     Qwen3-32B (think)  & 7761.6  & 5274.7 \\
%     Doubao-Seed-1.6       & 5288.1    & 3898.2 \\
%     \bottomrule
% \end{tabular}}
% \end{table}

\paragraph{Batch Inference Encourages Efficient Reasoning.} As discussed in the introduction, a major challenge for RLLMs is their tendency to overthink, producing unnecessarily long reasoning chains even for simple questions. To investigate whether batch inference can encourage more efficient reasoning, we conduct a series of controlled experiments. Specifically, we randomly select 500 samples from the DeepScaleR dataset~\citep{deepscaler2025} and evaluate several mainstream LLMs under two settings: (i) querying one question at a time (Vanilla), and (ii) querying two questions per prompt (Batch-2). As shown in Table~\ref{tab:batch_compression}, models including DeepSeek-R1~\citep{deepseekai2025deepseekr1incentivizingreasoningcapability}, Qwen3-32B (think)~\citep{yang2025qwen3technicalreport}, and Doubao-Seed-1.6~\citep{doubao1.6} consistently generate shorter outputs in the `Batch-2' setting, suggesting that batch inference naturally promotes more concise reasoning and that this effect generalizes well across different model architectures.
% \begin{figure}[t]
%     \centering
%     \includegraphics[width=0.6\textwidth]{figures/batch_vs_token_accuracy.pdf}
%     \caption{Impact of batch size on output length and accuracy using DeepSeek-R1.}
%     \label{fig:batch_vs_token_accuracy}
% \end{figure}
% \begin{figure}[t]
% \centering
% % ====== 左边子图 ======
% \begin{subfigure}{0.45\textwidth} % 子图宽度可调
% \centering
% % \renewcommand{\arraystretch}{1.2}
% \resizebox{\linewidth}{!}{ % 让表格适应子图宽度
% \begin{tabular}{lcc}
%     \toprule
%     \textbf{Model} & \textbf{Vanilla} & \textbf{Batch-2} \\
%     \midrule
%     Deepseek-R1 & 5640.4 & 4035.2 \\
%     Qwen3-32B (think) & 7761.6 & 5274.7 \\
%     Doubao-Seed-1.6 & 5288.1 & 3898.2 \\
%     \bottomrule
% \end{tabular}
% }
% \caption{Comparison of average output token lengths across different models under the `Vanilla' and `Batch-2' settings.}
% \label{fig:batch_table}
% \end{subfigure}
% \hfill
% % ====== 右边子图 ======
% \begin{subfigure}{0.45\textwidth}
% \centering
% \includegraphics[width=\linewidth]{figures/batch_vs_token_accuracy.pdf}
% \caption{Impact of batch size on output length and accuracy using DeepSeek-R1.}
% \label{fig:batch_curve}
% \end{subfigure}

% \caption{
% \ykw{Batch inference effects on reasoning efficiency.  
% (a) shows token length reduction across models when switching from single-question (Vanilla) to batch (Batch-2) inference.  
% (b) plots how increasing batch size affects both output length and accuracy for DeepSeek-R1, illustrating the \emph{resource competition pressure} phenomenon.}
% }
% \label{fig:batch_combined}
% \end{figure}

\begin{wraptable}{l}{0.45\textwidth} % r=靠右，宽度可调
\vspace{-10pt} % 如果顶部留白太大可以收缩一点
\caption{Comparison of average output token lengths across different models under the `Vanilla' and `Batch-2' settings.}
\label{tab:batch_compression}
\centering
\resizebox{\linewidth}{!}{ % 注意用 \linewidth 代替固定宽
\begin{tabular}{lcc}
    \toprule
    \textbf{Model} & \textbf{Vanilla} & \textbf{Batch-2} \\
    \midrule
    Deepseek-R1        & 5640.4 & 4035.2 \\
    Qwen3-32B (think)  & 7761.6 & 5274.7 \\
    Doubao-Seed-1.6    & 5288.1 & 3898.2 \\
    \bottomrule
\end{tabular}}
\vspace{-0.5em}
% \vspace{-12pt} % 如果底部留白太大可以收缩一点
\end{wraptable}

% 图
\begin{wrapfigure}{r}{0.45\textwidth} % 'r' 表示靠右
\centering
\vspace{-11em}
\includegraphics[width=\linewidth]{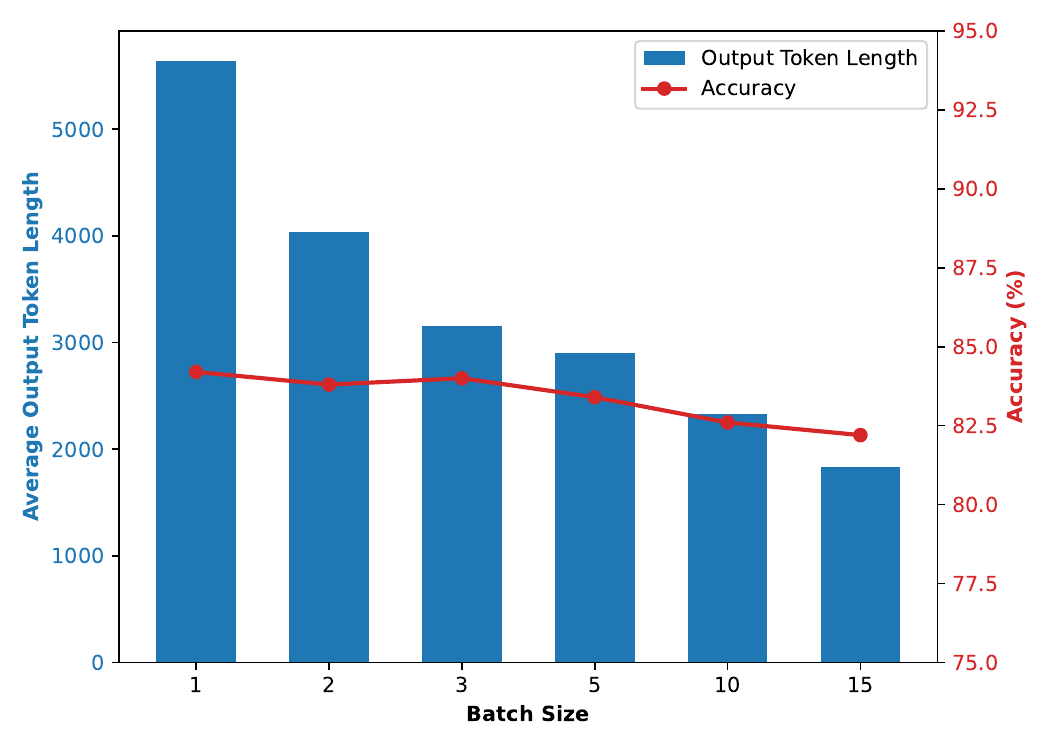}
\caption{Impact of batch size on output length and accuracy (DeepSeek-R1).}
\label{fig:batch_curve}
\vspace{-0.5em}
\end{wrapfigure}

\paragraph{Scaling Up Batch Size Further Enhances Efficiency.} To further analyze the effect, we vary the batch size using DeepSeek-R1 as a case study, testing batches of \textbf{2}, \textbf{3}, \textbf{5}, \textbf{10} and \textbf{15} questions. As shown in Figure~\ref{fig:batch_curve}, increasing the batch size leads to a continuous and substantial reduction in the average output length per question. Notably, this compression is achieved with only minimal degradation in answer accuracy, indicating that when more questions are packed into a single context window, the model tends to prioritize conciseness, allocating fewer tokens to simpler problems while preserving reasoning quality for more challenging ones. We refer to this emergent behavior as \emph{resource competition pressure}.

These findings provide compelling empirical evidence that RLLMs are capable of implicit reasoning compression when facing context constraints. The behavior of allocating reasoning resources based on task complexity, without any explicit instruction, points to a promising direction for mitigating the overthinking problem commonly observed in single-question inference. Building on this insight, our work is driven by a central research question: \textit{can we transfer the benefits of resource competition from batch inference to single-question settings?} If so, models could learn to reason adaptively, producing concise answers for simple queries while maintaining sufficient reasoning depth for more complex ones. To this end, we introduce Dynamic Reasoning Quota Allocation (DRQA), detailed in the following section.

\section{Methodology}
\label{sec:methodology}

Our goal is to enable RLLMs to assess question complexity and allocate reasoning resources adaptively, even when processing a single query. Ideally, the model should generate short responses for simple problems while preserving sufficient reasoning depth for more challenging ones, thereby improving inference efficiency without compromising answer accuracy. A key challenge in realizing this capability lies in how to effectively transfer ``resource competition pressure'' from batch inference to single-question settings. We first explore a straightforward solution via supervised fine-tuning (SFT) using batch-generated data. However, this approach revealed inherent limitations in teaching the model to internalize conciseness as a quality criterion. Inspired by recent advancements in Reinforcement Learning with Verifiable Rewards (RLVR) ~\citep{lambert2025tulu3pushingfrontiers,deepseekai2025deepseekr1incentivizingreasoningcapability}, we introduce Dynamic Reasoning Quota Allocation (DRQA), a reinforcement learning framework that explicitly encourages reasoning that is both accurate and concise. By optimizing an intrinsic reward aligned with these dual objectives, DRQA guides models to dynamically allocate reasoning resources, enabling more efficient and adaptive inference.

\begin{figure*}[t]
    \centering
    \includegraphics[width=\textwidth]{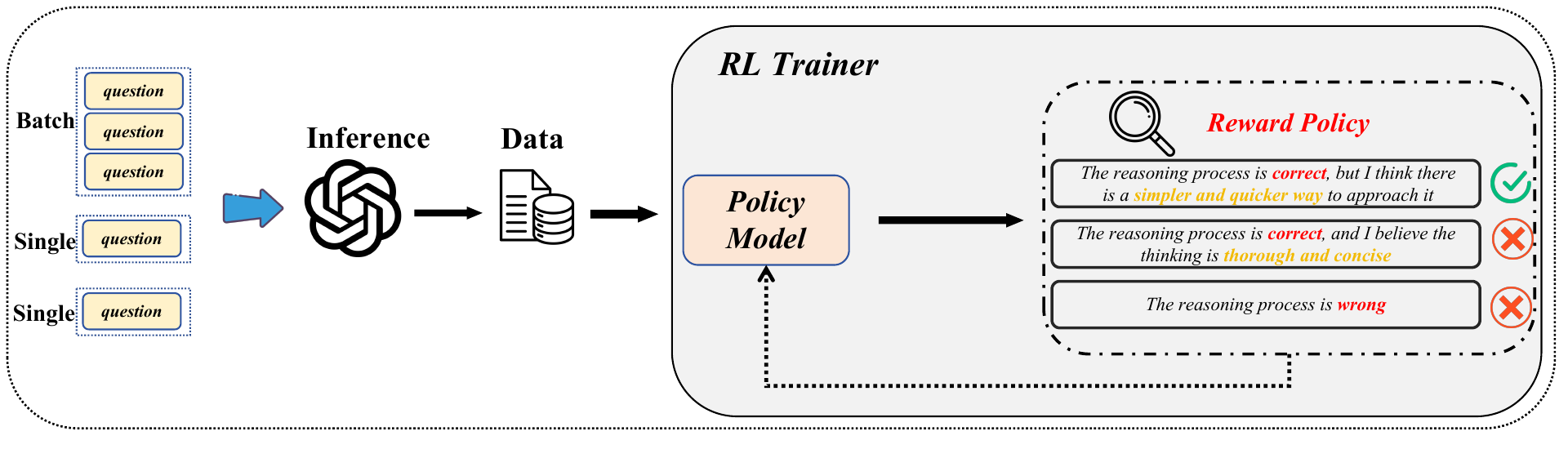}
    \caption{The pipeline of Dynamic Reasoning Quota Allocation (DRQA). Batched questions are input to LLM, producing reasoning chains labeled as A/B/C. Reinforcement learning trains the model to prefer concise and accurate reasoning for efficient resource allocation.}
    % \vspace{-1.2em}
    \label{fig:pipeline}
\end{figure*}

\subsection{Supervised Fine-Tuning with Batch Data}

\begin{table*}[t]
\centering
\renewcommand{\arraystretch}{1.2}
\caption{Single-question evaluation results of Qwen3-8B after SFT with data generated by batch inference. Batch-X denotes fine-tuning with data from batches of X questions, and Vanilla refers to the original model without SFT. }
\resizebox{\textwidth}{!}{
\begin{tabular}{lcccccccccccccc}
\toprule
\textbf{\textit{Method}} & \multicolumn{2}{c}{\textbf{GSM8K}} & \multicolumn{2}{c}{\textbf{Math500}} & \multicolumn{2}{c}{\textbf{AIME2024}} & \multicolumn{2}{c}{\textbf{GPQA-Diamond}} & \multicolumn{2}{c}{\textbf{AMC}} & \multicolumn{2}{c}{\textbf{AIME2025}} & \multicolumn{2}{c}{\textbf{Overall}} \\
                & \textbf{Acc}   & \textbf{tokens}   & \textbf{Acc}   & \textbf{tokens}   & \textbf{Acc}   & \textbf{tokens}   & \textbf{Acc}   & \textbf{tokens}   & \textbf{Acc}   & \textbf{tokens}   & \textbf{Acc}   & \textbf{tokens}   & \textbf{Acc}  & \textbf{tokens} \\
\midrule
Vanilla  & 95.67 & 1878.55  & 96.00 & 5270.58   & 74.67 & 15468.23 & 66.67 & 8685.21   & 97.50 & 8608.85   & 63.33 & 18058.65  & 82.31 & 9661.68 \\
Batch-2 & 96.67 & 575.64   & 95.00 & 2359.21   & 57.33 & 11100.55 & 53.54 & 6874.42   & 90.00 & 4136.58   & 45.33 & 13130.95  & 72.98 & 6362.89 \\
Batch-3 & 93.33 & 437.23   & 82.67 & 1593.53   & 26.00 & 5685.36  & 55.56 & 3555.65   & 77.50 & 4098.10   & 28.00 & 7400.53   & 60.51 & 3795.07 \\
Batch-5 & 93.33 & 336.81   & 69.67 & 434.50    & 9.33  & 2486.77  & 46.46 & 1190.23   & 42.50 & 922.25    & 7.33  & 2365.41   & 44.77 & 1289.33 \\
\bottomrule
\end{tabular}
}
\label{tab:sft_results}
\end{table*}

Our initial approach to transferring the benefits of resource competition into single-question inference is based on imitation learning, where we apply supervised fine-tuning (SFT) to mimic the efficient reasoning patterns exhibited by models during batch inference.

\paragraph{Method}
We use DeepSeek-R1~\citep{deepseekai2025deepseekr1incentivizingreasoningcapability} to perform batch inference over multiple questions sampled from DeepScaleR~\citep{deepscaler2025} and collect the generated responses, which are consistently more concise than those from single-question inference. Based on these results, we construct a dataset of ``question–concise answer'' pairs and apply full-parameter SFT on a Qwen3-8B~\citep{yang2025qwen3technicalreport}, with the goal of teaching it to generate similarly concise responses in single-question scenarios.

\paragraph{Experimental Results and Analysis}
We evaluate the fine-tuned models on a comprehensive set of reasoning benchmarks, including GSM8K~\citep{cobbe2021trainingverifierssolvemathgsm8k}, MATH-500~\citep{math500hendrycks2021measuringmathematicalproblemsolving}, AIME 2024~\citep{aime}, GPQA-Diamond~\citep{rein2023gpqagraduatelevelgoogleproofqa}, AMC~\citep{AMC2023}, and AIME 2025~\citep{aime}. The results shown in Table~\ref{tab:sft_results} indicate that SFT does lead to substantial reductions in output length. For example, on GSM8K, the average response length drops from 1878.55 to 575.64 tokens, a 69.36\% reduction, demonstrating that overthinking is mitigated to some extent. 

However, the efficiency gains come at a considerable cost to accuracy, particularly on more challenging tasks. As shown in Table~\ref{tab:sft_results}, Models fine-tuned with two-question batch data show a slight accuracy increase from 95.67\% to 96.67\% on GSM8K, while on MATH-500 accuracy drops from 96.00\% to 95.00\%, a decrease of 1.00\% compared to vanilla prompting. More notably, the performance degradation becomes increasingly severe with higher batch sizes and task complexity. On AIME 2024, accuracy falls from 74.67\% (Vanilla) to 57.33\% (Batch-2), 26.00\% (Batch-3), and just 9.33\% (Batch-5). These results suggest the emergence of catastrophic forgetting~\citep{luo2025empiricalstudycatastrophicforgetting}: in attempting to mimic the surface-level conciseness of batch responses, the model compromises its ability to perform the deeper, more nuanced reasoning necessary for solving complex problems.

In summary, while supervised fine-tuning with batch data effectively mitigates overthinking and improves inference efficiency, it comes at the cost of reasoning accuracy, especially on complex tasks, highlighting its limitations for real-world deployment. These shortcomings underscore the need for a more principled solution that can balance conciseness with reasoning depth, which motivates our proposed method: Dynamic Reasoning Quota Allocation (DRQA).

\subsection{Dynamic Reasoning Quota Allocation}
\label{sec:drqa}
Rather than imitating outputs from batch inference, we aim to endow the model with an intrinsic ability to evaluate and generate reasoning chains that are both accurate and concise. To this end, we propose Dynamic Reasoning Quota Allocation (DRQA), a reinforcement learning framework that enables RLLMs to dynamically allocate reasoning resources in single-question inference.

\paragraph{Core Idea}
The core idea of DRQA is to enhance the model's intrinsic reasoning capabilities by equipping it with the ability to evaluate the quality of its own reasoning chains. Specifically, the model is trained to make two key judgments: (i) whether a given reasoning chain is logically correct, and (ii) if correct, whether it is unnecessarily verbose. \emph{By developing this self-evaluation ability, the model learns to strike a balance between accuracy and conciseness during generation, effectively realizing adaptive resource allocation}.

\paragraph{Preference Data Construction}
\label{Training Data}
To train this evaluation ability, we construct a preference dataset consisting of multiple-choice question-answering samples. Each sample contains a question, a model-generated chain of thought (CoT), and three evaluation options that reflect different levels of reasoning quality:

\begin{itemize}[leftmargin=*,topsep=2pt,itemsep=2pt,parsep=0pt]
    \item \textbf{A}: The reasoning process is correct, but I think there is a simpler and quicker way to approach it.
    \item \textbf{B}: The reasoning process is correct, and I believe the thinking is thorough and concise.
    \item \textbf{C}: The reasoning process is wrong.
\end{itemize}

The dataset construction process involves three key steps. First, for ease of evaluation, we select all questions in the DeepScaleR~\citep{deepscaler2025} dataset whose answers are numbers of various types, resulting in approximately 30,000 samples. Second, for each question, we generate two types of reasoning chains using DeepSeek-R1~\citep{deepseekai2025deepseekr1incentivizingreasoningcapability}:
(1) \emph{vanilla CoTs} obtained by prompting the model with individual questions, and
(2) \emph{batch CoTs} generated by prompting the model with batched questions, followed by extracting the corresponding reasoning chain for each question. Finally, we assign labels based on reasoning correctness and conciseness: for vanilla CoTs, we label \textbf{A} if the reasoning is correct, and \textbf{C} if incorrect; for batch CoTs, we label \textbf{B} if the reasoning is correct, and \textbf{C} if incorrect. This labeling scheme enables the model to learn nuanced distinctions between correct-but-verbose reasoning (option A), correct-and-concise reasoning (option B), and incorrect reasoning (option C), thereby developing a clearer understanding of what constitutes a high-quality reasoning chain.

\paragraph{Reinforcement Learning Framework}
We use Group Relative Policy Optimization (GRPO)~\citep{shao2024deepseekmathpushinglimitsmathematical} to train the model to accurately classify each reasoning chain as A, B, or C, thus encouraging concise and accurate reasoning. Formally, the GRPO objective is defined as maximizing the likelihood of selecting the correct evaluation label:
\begin{equation}
\mathcal{L}_{\mathrm{GRPO}}(\theta) =
\mathbb{E}_{\tau \sim \mathcal{D}} \bigg[
    \sum_{a \in \mathcal{G}} 
        \log \pi_\theta(a \mid s) \; \hat{A}(a, s, a^*)
    - \beta\; \mathrm{KL} \!\left(\pi_\theta \,\|\, \pi_{\mathrm{old}}\right)
\bigg]
\end{equation}
where $\tau \sim \mathcal{D}$ denotes a sample from the dataset, with state $s$ representing the question, reasoning chain, and multiple-choice options (A, B, C); $a^*$ is the ground-truth label; $\mathcal{G} = \{A, B, C\}$ is the set of actions; $\hat{A}(a, s, a^*)$ is the relative advantage estimate, positive if $a = a^*$ and negative otherwise; $\text{KL}(\pi_\theta \| \pi_{\text{old}})$ is the KL divergence between the current and old policies, constrains the policy update; and $\beta$ is a regularization coefficient balancing learning efficiency and policy stability. This training objective encourages the model to assign higher probabilities to correct judgments while mitigating the risk of catastrophic forgetting caused by over-updating, a common issue encountered in SFT. As a result, the model gradually internalizes a preference for reasoning chains that are both correct and concise.

\paragraph{Summary}
DRQA enables the model to move beyond surface-level imitation and develop an intrinsic, reward-driven preference for high-quality reasoning. By balancing accuracy and conciseness, the model learns to allocate reasoning resources more effectively, addressing the limitations of SFT and supporting more efficient and adaptive inference in single-question settings.
\section{Experiments}
\label{sec:experiments}
In this section, we systematically evaluate the performance of the proposed DRQA algorithm, focusing on its ability to balance reasoning accuracy and efficiency. We compare DRQA against a range of strong baselines and provide an in-depth analysis of the results.

\subsection{Experimental Setup}
\label{sec:experimental_setup}
\paragraph{Models} We evaluate all methods using three widely adopted distilled models: DeepSeek-R1-Distill-Qwen-1.5B, DeepSeek-R1-Distill-Qwen-7B, and DeepSeek-R1-Distill-Llama-8B. All models are derived from the more powerful DeepSeek-R1~\citep{deepseekai2025deepseekr1incentivizingreasoningcapability} through large-scale distillation, offering a favorable trade-off between computational efficiency and reasoning capability.

\paragraph{Datasets}
For training, we use the dataset described in Section~\ref{Training Data}, constructed by performing batch inference with DeepSeek-R1 on the DeepScaleR~\citep{deepscaler2025} training set. This process yields over 50,000 multiple-choice examples annotated with reasoning quality labels.

\paragraph{Baselines}
To assess the effectiveness of DRQA, we compare it against a comprehensive set of strong baselines approaches (refer to Appendix~\ref{app:baselines} for detailed descriptions of the baselines). All baselines are either publicly released or carefully reproduced according to their original protocols.

\paragraph{Evaluation}

We evaluate the performance of different methods across a diverse set of benchmarks. For mathematical reasoning, we include GSM8K~\citep{cobbe2021trainingverifierssolvemathgsm8k}, MATH-500~\citep{math500hendrycks2021measuringmathematicalproblemsolving}, AIME 2024 and 2025~\citep{aime}, and AMC 2023~\citep{AMC2023}. For domain-specific scientific reasoning, we use the high-quality GPQA-diamond subset~\citep{rein2023gpqagraduatelevelgoogleproofqa}. Detailed descriptions of these datasets are provided in Appendix~\ref{app:datasets}. We use both accuracy and response length as evaluation metrics and report the average performance across all test sets. For the AIME datasets, which contain only 30 questions each, we repeatedly sample 5 responses for each
case and report the average results to ensure more stable and reliable evaluation.

All models are evaluated using a unified inference configuration to ensure fair comparison. Experiments are conducted with the vLLM framework on a computing cluster equipped with eight A800 (40GB) GPUs. The inference parameters are set to a temperature of $0.6$ and a maximum generation length of $32$K tokens.

\paragraph{Training Details}
\label{sec:training_details}
We use verl~\citep{sheng2024hybridflow} as the training framework. We set the batch size to 256, the number of rollouts to 16, the learning rate to $1 \times 10^{-6}$, and the maximum response length to 16K tokens. The model is trained for one epoch, consisting of 204 steps in total.

\subsection{Main Results}
\begin{table*}[htbp]
\centering
\renewcommand{\arraystretch}{1.15}
\caption{Performance of different methods using three RLLMs: DeepSeek-R1-Distill-Qwen-1.5B, DeepSeek-R1-Distill-Qwen-7B, and DeepSeek-R1-Distill-Llama-8B. DRQA achieves competitive or superior accuracy while greatly reducing token usage across all datasets and model variants, striking an excellent balance between performance and efficiency.}
\resizebox{\textwidth}{!}{
\begin{tabular}{llcccccccccccccc}
\toprule
\multirow{2}{*}{\textbf{\textit{Method}}} & \multicolumn{2}{c}{\textbf{GSM8K}} & \multicolumn{2}{c}{\textbf{MATH-500}} & \multicolumn{2}{c}{\textbf{AIME 2024}} & \multicolumn{2}{c}{\textbf{GPQA-Diamond}} &
\multicolumn{2}{c}{\textbf{AMC 2023}} & \multicolumn{2}{c}{\textbf{AIME 2025}} & \multicolumn{2}{c}{\textbf{Overall}} \\
& \textbf{Acc} & \textbf{Tokens} & \textbf{Acc} & \textbf{Tokens} & \textbf{Acc} & \textbf{Tokens} & \textbf{Acc} & \textbf{Tokens} & \textbf{Acc} & \textbf{Tokens} & \textbf{Acc} & \textbf{Tokens} & \textbf{Acc}$_{\text{All}}$ & \textbf{Tokens}$_{\text{All}}$ \\
\midrule
\rowcolor{gray!20}
\multicolumn{15}{l}{\textbf{DeepSeek-R1-Distill-Qwen-1.5B}} \\
\textit{Vanilla} & 84.67\% & 1928.96 & 83.33\% & 5536.14 & 28.67\% & 14394.61 & 30.84\% & 14731.59 & 72.50\% & 8830.10 & 23.67\% & 15323.3 & \multicolumn{1}{|l}{53.95\%} & 10124.12 \\
\textit{O1-Pruner} & 74.80\% & 458 & 82.20\% & 3212 & 28.90\% & 10361 & - & - & - & - & - & - & \multicolumn{1}{|c}{--} & -- \\
\textit{DAST} & 77.20\% & 586 & 83.00\% & 2428 & 26.90\% & 7745 & - & - & - & - & - & - & \multicolumn{1}{|c}{--} & -- \\
\textit{ShortBetter} & 63.67\% & \sota{107.86} & 60.33\% & \sota{1186.27} & 11.33\% & \sota{2935.68} & 21.72\% & \sota{1433.95} & 57.50\% & \sota{1260.43} & 12.67\% & \sota{3326.22} & \multicolumn{1}{|l}{\annotate{37.87\%}{blue}{-16.08}} & \annotate{\sota{1708.40}}{blue}{\textbf{-83.13\%}} \\
\textit{AdaptThink} & 86.00\% & 324.26 & 83.67\% & 1244.98 & 29.33\% & 7044.06 & 29.80\% & 4744.23 & 72.50\% & 2441.45 & 24.67\% & 7490.79 & \multicolumn{1}{|l}{\annotate{54.33\%}{red}{+0.38}} & \annotate{3881.63}{blue}{-61.66\%} \\
\textit{GRPO} & \sota{87.33\%} & 1691.19 & \sota{84.67\%} & 5743.01 & \sota{32.67\%} & 15017.54 & 27.78\% & 13809.53 & \sota{77.50\%} & 9378.21 & 24.00\% & 13082.98 & \multicolumn{1}{|l}{\annotate{55.66\%}{red}{+1.71}} & \annotate{9787.08}{blue}{-3.33\%} \\
\textit{GRPO+Length Penalty} & 86.00\% & 722.34 & 84.67\% & 2479.14 & 24.67\% & 9011.46 & 26.76\% & 6148.50 & 67.50\% & 3130.51 & 22.00\% & 9782.34 & \multicolumn{1}{|l}{\annotate{51.93\%}{blue}{-2.01}} & \annotate{5212.38}{blue}{-48.52\%} \\
\textit{SFT} & 81.67\% & 2296.54 & 80.33\% & 5465.95 & 25.33\% & 21337.44 & 27.27\% & 18540.94 & 65.00\% & 8806.48 & 19.33\% & 20258.82 & \multicolumn{1}{|l}{\annotate{49.82\%}{blue}{-4.13}} & \annotate{12784.36}{red}{+26.28\%} \\
\textit{DRQA(our)} & 86.67\% & 1427.63 & \sota{84.67\%} & 3488.08 & 32.00\% & 11008.31 & \sota{31.81\%} & 9148.83 & 75.00\% & 5355.03 & 24.00\% & 10382.12 & \multicolumn{1}{|l}{\annotate{\sota{55.69\%}}{red}{\textbf{+1.74}}} & \annotate{6801.67}{blue}{-32.82\%} \\
\midrule
\rowcolor{gray!20}
\multicolumn{15}{l}{\textbf{DeepSeek-R1-Distill-Qwen-7B}} \\
\textit{Vanilla} & 91.33\% & 1735.5 & 90.40\% & 5099.95 & 53.33\% & 13712.6 & 48.98\% & 13313.92 & 90.00\% & 6349.53 & 40.00\% & 14248.11 & \multicolumn{1}{|l}{69.01\%} & 9076.60 \\
\textit{DAST} & 86.70\% & 459 & 89.60\% & 2162 & 45.60\% & 7578 & - & - & - & - & - & - & \multicolumn{1}{|c}{--} & -- \\
\textit{O1-Pruner} & 87.60\% & 428 & 86.60\% & 2534 & 49.20\% & 9719 & - & - & - & - & - & - & \multicolumn{1}{|c}{--} & -- \\
\textit{Dynasor-CoT} & 89.60\% & 1285 & 89.00\% & 2971 & 46.70\% & 12695 & 30.50\% & 7639 & 85.00\% & 5980 & - & - & \multicolumn{1}{|c}{--} & -- \\
\textit{DEER} & 90.60\% & 6917 & 89.80\% & 2143 & 49.20\% & 9839 & 31.30\% & 5469 & 85.00\% & 4451 & - & - & \multicolumn{1}{|c}{--} & -- \\
\textit{ShortBetter} & 70.00\% & \sota{112.86} & 68.00\% & \sota{623.44} & 41.33\% & \sota{5005.96} & 43.43\% & \sota{1811.43} & 57.50\% & \sota{1567.50} & 30.67\% & 5393.96 & \multicolumn{1}{|l}{\annotate{51.82\%}{blue}{-17.19}} & \annotate{\sota{2419.19}}{blue}{\textbf{-73.35\%}} \\
\textit{AdaptThink} & 89.67\% & 296.94 & 91.67\% & 1839.59 & 54.00\% & 9894.05 & 51.52\% & 7128.95 & 87.50\% & 3287.95 & 39.33\% & 12454.59 & \multicolumn{1}{|l}{\annotate{68.95\%}{blue}{-0.06}} & \annotate{5817.01}{blue}{-35.91\%} \\
\textit{AutoL2S} & 93.33\% & 444.8 & 83.33\% & 3113.93 & 40.67\% & 6499.32 & 45.39\% & 2553.01 & 85.00\% & 2613.05 & 31.33\% & \sota{3669.53} & \multicolumn{1}{|l}{\annotate{63.18\%}{blue}{-5.84}} & \annotate{3148.94}{blue}{-65.31\%} \\

\textit{GRPO} & \sota{93.67\%} & 1524.24 & \sota{92.00\%} & 4532.21 & \sota{54.67\%} & 12013.92 & 47.47\% & 12124.10 & 87.50\% & 5130.13 & \sota{41.33\%} & 12192.12 & \multicolumn{1}{|l}{\annotate{69.44\%}{red}{+0.43}} & \annotate{7919.45}{blue}{-12.75\%} \\

\textit{GRPO+Length Penalty} & 91.33\% & 876.25 & 91.33\% & 2751.13 & 52.00\% & 7213.11 & 45.96\% & 7124 & \sota{92.50\%} & 3256.02 & 39.67\% & 6058.40 & \multicolumn{1}{|l}{\annotate{68.80\%}{blue}{-0.21}} & \annotate{4546.49}{blue}{-49.91\%} \\

\textit{SFT} & 92.33\% & 1317.85 & 92.00\% & 3824.43 & 44.67\% & 14903.82 & 46.97\% & 12385.43 & 77.50\% & 5519.55 & 32.00\% & 13931.80 & \multicolumn{1}{|l}{\annotate{64.25\%}{blue}{-4.76}} & \annotate{8647.15}{blue}{-4.73\%} \\

\textit{DRQA(our)} & 92.67\% & 1324.24 & 91.40\% & 3902.74 & \sota{54.67\%} & 10007.18 & \sota{49.50\%} & 8988.50 & \sota{92.50\%} & 4463.03 & 40.67\% & 9545.44 & \multicolumn{1}{|l}{\annotate{\sota{70.24\%}}{red}{\sota{+1.23}}} & \annotate{6371.85}{blue}{-29.80\%} \\
\midrule
\rowcolor{gray!20}
\multicolumn{15}{l}{\textbf{DeepSeek-R1-Distill-Llama-8B}} \\
\textit{Vanilla} & 91.67\% & 1829.12 & 90.00\% & 5417.41 & 49.33\% & 13585.12 & 48.98\% & 11845.27 & 87.50\% & 7177.73 & 38.67\% & 14260.26 & \multicolumn{1}{|l}{67.69\%} & 9019.15 \\
\textit{GRPO} & 92.33\% & 1605.94 & \sota{91.67\%} & 4812.02 & 50.67\% & 12897.09 & 46.46\% & 9869.20 & 90.00\% & 7600.58 & 39.33\% & 12204.58 & \multicolumn{1}{|l}{ \annotate{68.41\%}{red}{+0.72}} & \annotate{8164.90}{blue}{-9.47\%}  \\
\textit{GRPO+Length Penalty} & 91.67\% & \sota{875.66} & 91.33\% & \sota{2753.43} & 48.00\% & \sota{7192.28} & 45.96\% & \sota{7055.54} & 90.00\% & \sota{3236.22} & 38.00\% & \sota{8040.74} & \multicolumn{1}{|l}{ \annotate{67.49\%}{blue}{-0.20}} & \sota{ \annotate{4858.98}{blue}{-46.13\%}} \\
\textit{SFT} & 90.67\% & 1315.83 & 90.00\% & 3825.52 & 44.67\% & 14881.25 & 44.95\% & 10897.06 & 75.00\% & 5509.82 & 32.67\% & 13915.29 & \multicolumn{1}{|l}{\annotate{62.99\%}{blue}{-4.70}} & \annotate{8390.80}{blue}{-6.97\%} \\
\textit{DRQA(our)} & \sota{93.00\%} & 1594.70 & 91.33\% & 4180.83 & \sota{50.67\%} & 9940.46 & \sota{50.00\%} & 8986.63 & \sota{92.50\%} & 4463.43 & \sota{39.33\%} & 9542.11 & \multicolumn{1}{|l}{\sota{\annotate{69.47\%}{red}{+1.78}} }& \annotate{6451.36}{blue}{-28.47\%} \\

\bottomrule
\end{tabular}
}
\label{tab:main_results}
\end{table*}

As shown in Table~\ref{tab:main_results}, DRQA demonstrates clear superiority in both answer accuracy and response efficiency across all mathematical benchmarks. For example, on GSM8K with the 1.5B model, DRQA achieves an accuracy of 86.67\%, outperforming the vanilla baseline by 2 percentage points, while reducing average token usage from 1928.96 to 1427.63, a 25.9\% reduction. Similar patterns are observed on more challenging datasets such as AIME 2024 and MATH-500, where DRQA maintains high accuracy while significantly reducing output length. These results highlight DRQA’s effectiveness in dynamically allocating reasoning resources, enabling it to strike a favorable balance between accuracy and efficiency across tasks of varying difficulties. Moreover, DRQA demonstrates strong generalization on out-of-distribution (OOD) benchmarks, as evidenced by its performance on GPQA-Diamond.

We also compare DRQA with aggressive compression methods such as ShorterBetter~\citep{yi2025shorterbetterguidingreasoningmodels} and DAST~\citep{shen2025dastdifficultyadaptiveslowthinkinglarge}, which can reduce output length even further, for example, generating outputs as short as 107.86 tokens on GSM8K. However, these methods often suffer from severe accuracy degradation, with performance drops exceeding 20 percentage points in some cases. This highlights a key limitation of methods that rely solely on length-based reward signals: they tend to compromise the logical integrity of reasoning chains, limiting their practical applicability.

Notably, DRQA remains highly effective on larger models. On GSM8K with the 7B model, DRQA improves accuracy by 1.34\% over the baseline while reducing token usage by 23.6\%. On Llama-8B, DRQA achieves a 1.78\% accuracy gain while cutting token usage by 28.47\%, highlighting its ability to enhance performance and efficiency at larger model scales. Across all benchmarks, it consistently achieves the most favorable trade-off between accuracy and output efficiency. Compared to strong baselines such as DAST~\citep{shen2025dastdifficultyadaptiveslowthinkinglarge}, O1-Pruner~\citep{luo2025o1prunerlengthharmonizingfinetuningo1like}, Dynasor-CoT~\citep{fu2025reasoningdynasor}, and DEER~\citep{xia2024deerdelayresilientframeworkreinforcement}, DRQA not only matches or surpasses them in length reduction but, more importantly, maintains state-of-the-art reasoning accuracy.

Overall, DRQA achieves an average accuracy improvement of 1.58 percentage points and an average token usage reduction of 30.4\% across all evaluated benchmarks and all three model variants. These results provide compelling evidence that DRQA effectively transfers the benefits of ``resource competition pressure'' from batch inference to single-question settings, establishing a strong foundation for the efficient and scalable deployment of RLLMs.

\subsection{Generalization to Code Generation}
We further assess DRQA on the LiveCodeBench benchmark~\citep{jain2024livecodebenchholisticcontaminationfree}, a contamination-free suite of code-related tasks collected from competitive programming platforms. Our evaluation uses 342 newly released Python problems spanning September~2024 and April~2025. As shown in Table~\ref{tab:livecode}, DRQA consistently reduces token usage by about 23\%--29\% across all three model sizes, while also improving accuracy. For example, on \textit{DeepSeek-R1-Distill-Qwen-7B}, DRQA shortens outputs from 8724.27 to 6648.77 tokens (-23.79\%) and improves accuracy by 1.75\%, demonstrating its strong generalizability to the code generation domain.

\begin{table}[t]
\centering
\renewcommand{\arraystretch}{1.0}
\caption{
Performance on LiveCodeBench. DRQA consistently reduces token usage while improving accuracy across all model sizes.
}
\resizebox{\textwidth}{!}{
\begin{tabular}{lcccccc}
\toprule
\multirow{2}{*}{\textbf{Method}} 
    & \multicolumn{2}{c}{\textbf{DeepSeek-R1-Distill-Qwen-1.5B}} 
    & \multicolumn{2}{c}{\textbf{DeepSeek-R1-Distill-Qwen-7B}} 
    & \multicolumn{2}{c}{\textbf{DeepSeek-R1-Distill-Llama-8B}}\\
 & \textbf{Acc} & \textbf{Tokens} 
 & \textbf{Acc} & \textbf{Tokens} 
 & \textbf{Acc} & \textbf{Tokens} \\
\midrule
\textit{Vanilla} 
& 13.16\% & 11261.72 
& 30.70\% & 8724.27 
& 31.87\% & 9012.31 \\
\textit{DRQA(our)}
& \textbf{13.74\%} & \textbf{8124.20} 
& \textbf{32.45\%} & \textbf{6648.77} 
& \textbf{32.33\%} & \textbf{6426.68} \\
\bottomrule
\end{tabular}
}
\label{tab:livecode}
% \vspace{-2em}
\end{table}

\subsection{Ablation Study}
\begin{table*}[t]
\caption{Ablation experiments across different training paradigms.}
\centering
\renewcommand{\arraystretch}{1.0}
\resizebox{\textwidth}{!}{
\begin{tabular}{lcccccccccc}
\toprule
\multirow{2}{*}{\textbf{Method}} & \multicolumn{2}{c}{\textbf{GSM8K}} & \multicolumn{2}{c}{\textbf{MATH-500}} & \multicolumn{2}{c}{\textbf{AIME 2024}} & \multicolumn{2}{c}{\textbf{Overall}} \\
& \textbf{Acc} & \textbf{tokens} & \textbf{Acc} & \textbf{tokens} & \textbf{Acc} & \textbf{tokens} & \textbf{Acc} & \textbf{tokens} \\
\midrule
\textit{Vanilla} & 91.33\% & 1735.5 & 90.40\% & 5099.95 & 53.33\% & 13712.6 & \multicolumn{1}{|l}{78.35\%} & 6849.35 \\

\textit{DRQA (Batch-2)} & 92.67\% & 1324.24 & 91.33\% & 3902.74 & 54.67\% & 10007.18 &  \multicolumn{1}{|l}{\annotate{\sota{79.58\%}}{red}{\sota{+1.23}}} & \annotate{5078.05}{blue}{-25.86\%}  \\

\textit{DRQA (Batch-3)} & 91.67\% & 1212.59 & 90.20\% & 3311.20 & 53.33\% & 8805.24  &  \multicolumn{1}{|l}{\annotate{78.40\%}{red}{+0.05}} & \annotate{4443.01}{blue}{-35.13\%}  \\

\textit{DRQA (Batch-5)} & 90.67\% & 1158.88 & 89.80\% & 2675.81 & 49.33\% & 7366.80&  \multicolumn{1}{|l}{\annotate{76.60\%}{blue}{-1.75}} & \annotate{\sota{ 3733.83}}{blue}{\sota{-45.49\%}}  \\
\textit{Qwen2.5-7B Data + RL} & 90.00\% & 1434.65 & 89.60\% & 3313.12 & 50.67\% & 12190.59 &  \multicolumn{1}{|l}{\annotate{76.76\%}{blue}{-1.60}} & \annotate{ 5646.12}{blue}{-17.57\%}  \\

\textit{Batch-2 Data + CFT} & 89.67\% & 1361.00 & 88.20\% & 3973.54 & 49.66\% & 10012.55 & \multicolumn{1}{|l}{\annotate{75.84\%}{blue}{-2.51}} & \annotate{ 5115.70}{blue}{-25.31\%}  \\
\bottomrule
\end{tabular}
}
\label{tab:ablation}
% \vspace{-2em}
\end{table*}

To thoroughly assess the contribution of each core component in DRQA, we conduct a series of ablation studies that isolate the effects of different training paradigms and input conciseness on reasoning performance and efficiency. All experiments are performed using the same benchmark datasets, evaluation metrics, and base model (DeepSeek-R1-Distill-Qwen-7B) as in the main study, with consistent inference configurations to ensure fair comparison.

\paragraph{Effect of Batch Size in DRQA Data Construction.} We investigate the impact of different batch sizes on model performance. Specifically, we construct preference datasets by prompting DeepSeek-R1 with batches of 2, 3, or 5 questions, then splitting the outputs into individual reasoning chains for downstream RL training. This design allows us to analyze how increasing levels of resource competition influence both answer accuracy and response efficiency within the DRQA framework.

\paragraph{Replacing Batch Reasoning Data with Qwen2.5-7B Concise Chains}

To evaluate the importance of batch-induced resource competition, we consider an alternative setting where the preference dataset is constructed using concise reasoning chains generated directly by Qwen2.5-7B~\citep{qwen2025qwen25technicalreport}, without leveraging batch inference. This comparison allows us to disentangle the effects of resource-driven compression from those achieved solely through the model's inherent ability to generate concise outputs.

\paragraph{Critique Fine-Tuning with Preference data} 

Beyond reinforcement learning, we also evaluate the Critique Fine-Tuning (CFT) paradigm~\citep{wang2025critiquefinetuninglearningcritique} as an alternative training strategy, applying it to the preference data we constructed.

\subsubsection{Results and Analysis}
\begin{wrapfigure}{r}{0.5\linewidth} % 'r' 表示靠右
\centering
\vspace{-2em}
\includegraphics[width=\linewidth]{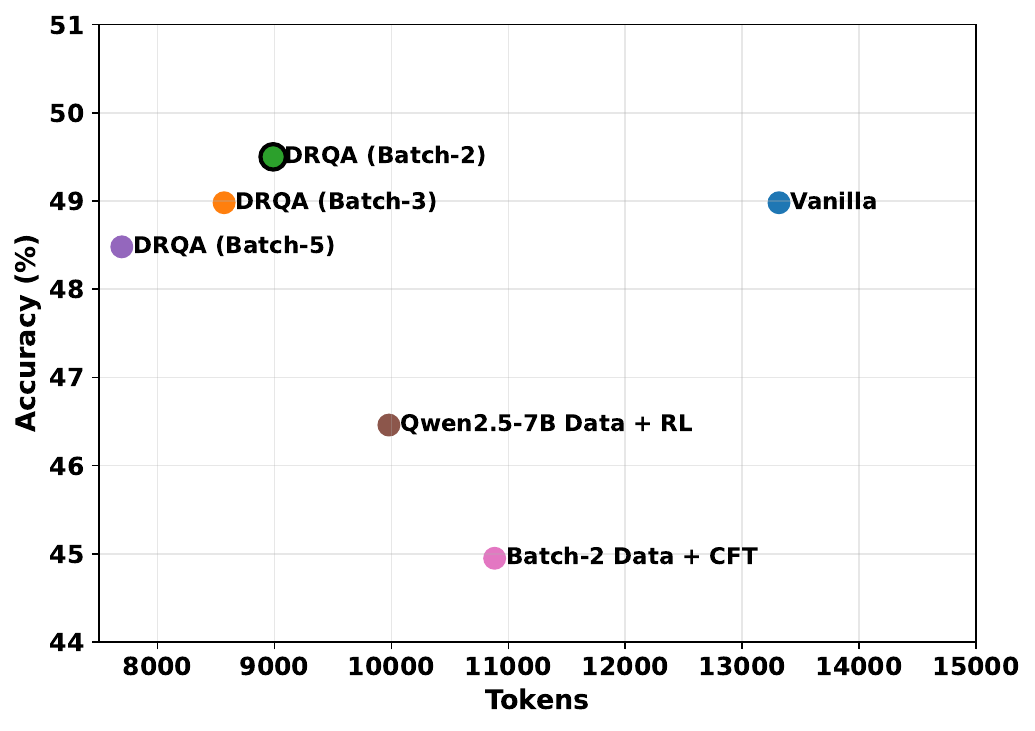}
\caption{The efficiency-accuracy trade-off on GPQA-diamond for DRQA and ablation variants.}
\label{fig:gpqa_ood_ablation}
\vspace{-2em}
\end{wrapfigure}

% \begin{figure}[htbp]
%     \centering
%     \includegraphics[width=0.6\linewidth]{figures/ablation_ood.pdf}
%     \caption{
%     The efficiency-accuracy trade-off on GPQA-diamond for DRQA and ablation variants.
%     }
%     \label{fig:gpqa_ood_ablation}
% \end{figure}

Table~\ref{tab:ablation} presents the results of our ablation study. As batch size increases, the model produces increasingly concise outputs, with token usage reduced by up to 45\% for larger batches. However, this efficiency gain comes at the cost of declining accuracy, highlighting a trade-off between efficiency and correctness. Notably, a batch size of 2 achieves the best balance, improving accuracy while significantly reducing token consumption compared to the vanilla baseline.

When compared to concise reasoning chains generated directly by Qwen2.5-7B~\cite{qwen2025qwen25technicalreport} without batch inference, we observe that only batch-induced compression achieves both high efficiency and strong accuracy. Similarly, while Critique Fine-Tuning helps reduce output length, it leads to a notable accuracy drop, underscoring the importance of reinforcement learning for preserving reasoning quality. Figure~\ref{fig:gpqa_ood_ablation} further supports these insights, showing that DRQA achieves the best overall trade-off on the OOD dataset GPQA-Diamond, highlighting its robustness across both in-distribution and out-of-distribution scenarios.

\section{Related Work}

\subsection{Reasoning Large Language Models}

Recent advances in reasoning large language models (RLLMs), such as OpenAI-O3~\citep{openaio3}, Deepseek-R1~\citep{deepseekai2025deepseekr1incentivizingreasoningcapability}, and QwQ~\citep{qwq32b} leverage chain-of-thought~\citep{wei2023chainofthoughtpromptingelicitsreasoning} for step-by-step reasoning, achieving state-of-the-art performance across tasks including mathematical reasoning, coding, and complex question answering. CoT allows these models to leverage inference-time scaling by generating multiple reasoning steps that explore alternative solution paths, thereby significantly enhancing accuracy over single-pass generation. To further improve correctness, a variety of methods have been proposed, including self-consistency~\citep{wang2023selfconsistencyimproveschainthought}, beam search~\citep{yao2023treethoughtsdeliberateproblem}, and reinforcement learning-based post-training~\citep{deepseekai2025deepseekr1incentivizingreasoningcapability}, which encourage iterative self-reflection and help reduce logical errors. Additional search-based approaches, such as Monte Carlo Tree Search (MCTS)~\citep{gao2024interpretablecontrastivemontecarlo}, have been employed to expand the scope of exploration in complex problem-solving scenarios.
Our work focuses on further improving the efficiency of such reasoning models.

\subsection{Efficient Reasoning}

Reasoning efficiency in RLLMs~\citep{qu2025surveyefficientreasoninglarge,sui2025stopoverthinkingsurveyefficient} refers to balancing task quality and computational cost. Models like OpenAI-O3~\citep{openaio3} and DeepSeek-R1~\citep{deepseekai2025deepseekr1incentivizingreasoningcapability} often generate too long and redundant reasoning chains, over explaining simple problems while sometimes offering shallow reasoning for complex ones. Main approaches for improving efficiency include:
\begin{itemize}
[leftmargin=*,topsep=2pt,itemsep=2pt,parsep=0pt]
\item \textbf{Inference time control}: Methods such as TALE~\citep{han2025tokenbudgetawarellmreasoning}, DEER~\citep{yang2025dynamicearlyexitreasoning}
apply token budgets or early exit strategies inspired by dual-system theory.
\item \textbf{Chain compression and supervised tuning}: TokenSkip~\citep{xia2025tokenskipcontrollablechainofthoughtcompression}, CoT-Valve~\citep{ma2025cotvalvelengthcompressiblechainofthoughttuning}, and AutoL2S~\citep{luo2025autol2sautolongshortreasoning} use supervised fine-tuning or distillation to shorten reasoning chains, often improving conciseness but sometimes at the expense of complex reasoning.
\item \textbf{Reinforcement learning approaches}: DAST~\citep{shen2025dastdifficultyadaptiveslowthinkinglarge}, O1-Pruner~\citep{luo2025o1prunerlengthharmonizingfinetuningo1like}, and S-GRPO~\citep{dai2025sgrpoearlyexitreinforcement} introduce reward functions to penalize lengthy outputs and promote token efficiency, supporting adaptive reasoning with little loss of accuracy.
\end{itemize}
These methods largely depend on fixed budgets or hand crafted rewards. Our DRQA instead transfers the ``resource competition pressure' observed in batch inference to single-question settings, enabling models to automatically adjust reasoning length according to problem complexity, providing brief responses for simple questions and detailed explanations for challenging ones without manual constraints.
\section{Conclusion}

This paper introduces Dynamic Reasoning Quota Allocation (DRQA), a novel approach aimed at addressing the overthinking problem in reasoning large language models (RLLMs). Motivated by the observation that resource competition pressure in batch inference naturally encourages efficient reasoning, DRQA leverages batch-generated data and reinforcement learning to transfer the benefits of resource competition from batch inference to single-question scenarios. Specifically, the model is trained to develop an internal preference for reasoning processes that balance conciseness with accuracy, allowing it to produce short answers for straightforward questions while preserving adequate reasoning depth when tackling more complex ones. Extensive experimental results and analysis show that DRQA significantly reduces token consumption while maintaining, or even improving, accuracy. By effectively alleviating overthinking, DRQA offers a new direction for more efficient and scalable deployment of RLLMs.

\bibliography{iclr2026_conference}
\bibliographystyle{iclr2026_conference}

\appendix
\newpage
% \onecolumn
\appendix
% \section{Full Results for SFT}
% \label{app:sft_results}

% Table~\ref{tab:sft_results} reports the complete single-question evaluation results of Qwen3-8B after SFT with data generated by batch inference (Batch-N). As discussed in Section~\ref{sec:methodology}, larger batch sizes lead to stronger token reduction but more pronounced accuracy drops, especially on complex reasoning tasks.

\section{Baseline Methods}
\label{app:baselines}
We consider the following baseline methods in our experiments:
\begin{itemize}[leftmargin=*,topsep=2pt,itemsep=2pt,parsep=0pt]
    \item \textbf{GRPO:} We train a model on the DeepScaleR~\citep{deepscaler2025} dataset using the Group Relative Policy Optimization algorithm, where only answer correctness is used as the reward signal.
    \item \textbf{GRPO+Length Penalty:} This variant further introduces a length penalty to the reward design: for correct answers, shorter responses yield higher rewards, while for incorrect answers, longer responses incur greater penalties. This encourages the model to produce concise and accurate reasoning.
    \item \textbf{SFT (Supervised Fine-Tuning):} We perform full-parameter supervised fine-tuning on the model using question-answer pairs generated via batch inference of Deepseek-R1 on the DeepScaleR~\citep{deepscaler2025} dataset.
    \item \textbf{AdaptThink}~\citep{zhang2025adaptthinkreasoningmodelslearn}: This approach encourages adaptive selection between direct answer and step-by-step reasoning (Chain-of-Thought) based on question difficulty. Training objectives and sample balancing enable the model to flexibly explore both thinking modes, improving reasoning efficiency and performance.
    \item \textbf{AutoL2S}~\citep{luo2025autol2sautolongshortreasoning}: A dynamic, model-agnostic framework that annotates each question with both long and short Chain-of-Thought (CoT) solutions. By marking simple questions with \texttt{<EASY>}, the model is trained to automatically select concise CoT for simple problems and detailed reasoning for complex ones.
    \item \textbf{DAST}~\citep{shen2025dastdifficultyadaptiveslowthinkinglarge}: DAST explicitly quantifies problem difficulty via a token length budget and employs a reward that penalizes redundant reasoning on simple problems while encouraging extensive CoT for difficult ones. This preference data is optimized via SimPO, enabling efficient dynamic control over reasoning path length.
    \item \textbf{O1-Pruner}~\cite{luo2025o1prunerlengthharmonizingfinetuningo1like}: Based on reinforcement learning, this method rewards shorter CoT traces without compromising accuracy. It employs an offline PPO-like procedure to prune redundant reasoning while preserving or even improving correctness.
    \item \textbf{ShorterBetter}~\citep{yi2025shorterbetterguidingreasoningmodels}: This RL-based approach defines the optimal length for each question as the shortest possible correct response and leverages this dynamic signal as a reward for GRPO-based training, guiding the model toward concise yet accurate answers.
    \item \textbf{Dynasor-CoT}~\citep{fu2025reasoningdynasor}: Without extra training, this method dynamically truncates reasoning by probing intermediate answers, monitoring consistency, and detecting hesitancy tokens. This yields substantial token savings while preserving accuracy.
    \item \textbf{DEER}~\citep{xia2024deerdelayresilientframeworkreinforcement}: DEER employs a dynamic early-exit mechanism by monitoring reasoning transitions (such as ``Wait'') to induce trial answers. Decisions to terminate CoT generation are based on confidence estimation, reducing reasoning length without additional training.
\end{itemize}

All baseline models are tested under identical inference configurations and on the same benchmark datasets to guarantee fair and reliable comparison. For each baseline, we use either the officially released model or reproduce the method using released data and code.

\section{Dataset Details}
\label{app:datasets}

\paragraph{Mathematical Reasoning Datasets}
\begin{itemize}[leftmargin=*,topsep=2pt,itemsep=2pt,parsep=0pt]
    \item \textbf{GSM8K}: This dataset contains 8,500 English elementary school single-step math reasoning questions. It serves as one of the mainstream benchmarks for evaluating the math reasoning abilities of large language models, focusing on basic arithmetic reasoning skills.
    \item \textbf{MATH-500}: Includes 500 medium-difficulty mathematical problems covering algebra, geometry, number theory, and other areas, designed to test the model's comprehensive mathematical reasoning ability.
    \item \textbf{AIME 2024/2025}: Originating from the American Invitational Mathematics Examination 2024 and 2025, each set contains 30 high-difficulty math questions, mainly assessing complex mathematical reasoning and problem-solving skills.
    \item \textbf{AMC 2023}: 40 questions from the American Mathematics Competitions (AMC), covering middle to high school levels, examining fundamental and advanced mathematics knowledge and problem solving abilities.
\end{itemize}

\paragraph{Scientific Reasoning Dataset}
To evaluate model reasoning performance in other domains, we use the high-quality GPQA-diamond subset from the GPQA dataset. GPQA-diamond is a refined version of GPQA, focusing on challenging, high-quality scientific domain questions and designed to provide a comprehensive assessment of scientific understanding and reasoning ability.

\section{Prompt Template}

\begin{tcolorbox}[title=Prompt for Batch Inference, label={fig:batch_inference}]
Please answer the following math problems in order and summarize all answers at the end:
Your response should be in the following format: 
\begin{verbatim}
[Solution Process]  
Provide a detailed solution for each problem...

[Final Answer]  
1. \\boxed{{Answer1}}  
2. \\boxed{{Answer2}}  
...  
n. \\boxed{{Answern}}  
\end{verbatim}
Below is the list of questions: 
\begin{verbatim}
{numbered_questions}
\end{verbatim}
\end{tcolorbox}

\begin{tcolorbox}[title=Prompt for Evaluation, label={fig:evaluation}]
\begin{verbatim}
{origin_question}\n\n
Please reason step by step, and put your final answer 
within \\boxed{}.
\end{verbatim}

\end{tcolorbox}

\end{document}